\documentclass[sigconf,screen,nonacm]{acmart}
\usepackage{multicol}
\usepackage{multirow}
\usepackage{algorithm}
\usepackage{algorithmic}
\newcommand{\el}{$\mathcal{EL}^{++}$}
\usepackage{color}
\usepackage{todonotes}
\usepackage{amsthm}

\AtBeginDocument{%
  \providecommand\BibTeX{{%
    \normalfont B\kern-0.5em{\scshape i\kern-0.25em b}\kern-0.8em\TeX}}}
\makeatletter
\newenvironment{breakablealgorithm}
  {
   \begin{center}
     \refstepcounter{algorithm}
     \hrule height.8pt depth0pt \kern2pt
     \renewcommand{\caption}[2][\relax]{
       {\raggedright\textbf{\ALG@name~\thealgorithm} ##2\par}%
       \ifx\relax##1\relax 
         \addcontentsline{loa}{algorithm}{\protect\numberline{\thealgorithm}##2}%
       \else 
         \addcontentsline{loa}{algorithm}{\protect\numberline{\thealgorithm}##1}%
       \fi
       \kern2pt\hrule\kern2pt
     }
  }{
     \kern2pt\hrule\relax
   \end{center}
  }

\setcopyright{acmcopyright}
\copyrightyear{2023}
\acmYear{2023}

\acmConference[WSDM '23]{WSDM '23: The 16th International Conference on Web Search and Data Mining}{February 27 - March 3, 2022}{Singapore}
\acmBooktitle{WSDM '23: The 16th International Conference on Web Search and Data Mining, February 27 - March 3, 2022, Singapore}

\begin{document}

\title{TAR: Neural Logical Reasoning across TBox and ABox}

\author{Zhenwei Tang} \affiliation{%
\institution{KAUST} \city{Thuwal} \country{Saudi Arabia}}
\email{zhenwei.tang@kaust.edu.sa}

\author{Shichao Pei} \affiliation{%
\institution{University of Notre Dame} \city{Indiana} \country{USA}}
\email{spei2@nd.edu}

\author{Xi Peng} \affiliation{%
\institution{KAUST} \city{Thuwal} \country{Saudi Arabia}}
\email{xi.peng@kaust.edu.sa}

\author{Fuzhen Zhuang} \affiliation{%
\institution{Beihang University} \city{Beijing} \country{China}}
\email{zhuangfuzhen@buaa.edu.cn}

\author{Xiangliang Zhang} \affiliation{%
\institution{University of Notre Dame} \city{Indiana} \country{USA}}
\email{xzhang33@nd.edu}

\author{Robert Hoehndorf} \affiliation{%
\institution{KAUST} \city{Thuwal} \country{Saudi Arabia}}
\email{robert.hoehndorf@kaust.edu.sa}


\begin{abstract}
Many ontologies, i.e., Description Logic (DL) knowledge bases, have been developed to provide rich knowledge about various domains.
An ontology consists of an ABox, i.e., assertion axioms between two entities or between a concept and an entity, and a TBox, i.e., terminology axioms between two concepts.
Neural logical reasoning (NLR) is a fundamental task to explore such knowledge bases, which aims at answering multi-hop queries with logical operations based on distributed representations of queries and answers. 
While previous NLR methods can give specific \textbf{entity-level} answers, i.e., ABox answers, they are not able to provide descriptive \textbf{concept-level} answers, i.e., TBox answers, where each concept is a description of a set of entities. 
In other words, previous NLR methods only reason over the ABox of an ontology while ignoring the TBox.
In particular, providing TBox answers enables inferring the explanations of each query with descriptive concepts, which make answers comprehensible to users and are of great usefulness in the field of applied ontology. 
In this work, we formulate the problem of \underline{n}eural
\underline{l}ogical \underline{r}easoning across \textbf{\underline{T}Box} and
\textbf{\underline{A}Box} (TA-NLR), solving which needs to address challenges in \textbf{incorporating}, \textbf{representing}, and \textbf{operating on concepts}. We propose an original solution named TAR for TA-NLR. Firstly, we incorporate description logic based ontological axioms to provide the source of concepts. Then, we represent concepts and queries as fuzzy sets, i.e., sets whose elements have degrees of membership, to bridge concepts and queries with entities. Moreover, we design operators involving concepts on top of fuzzy set representation of concepts and queries for optimization and inference.  Extensive experimental results on two real-world datasets demonstrate the effectiveness of TAR for TA-NLR.
\end{abstract}
\keywords{Neural logical reasoning, Knowledge representation
  learning.}

\maketitle

\section{Introduction}
\begin{figure}[h!]
  \centering
  \includegraphics[width=0.9\linewidth]{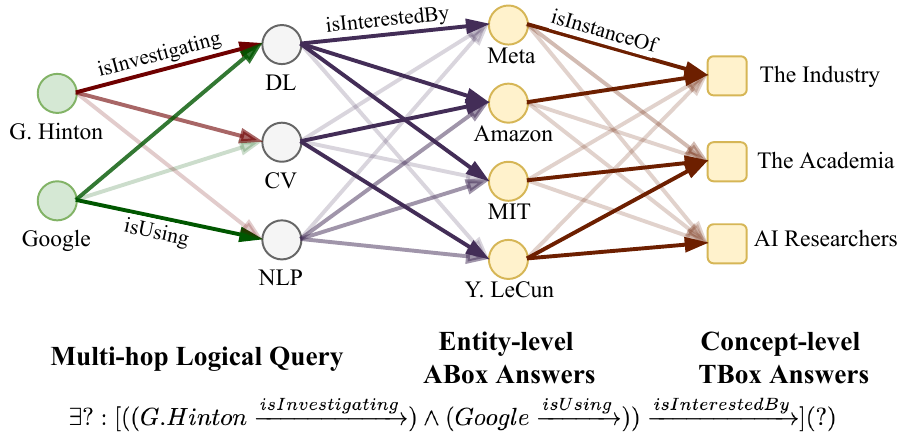}
  \caption{An example of TA-NLR. The query is
``who will be interested in techniques that G. Hinton is investigating
and Google is using?''. The answers are not only entity-level ABox answers as yellow circles: \textit{Meta}, \textit{Amazon}, \textit{MIT}, and \textit{Y. LeCun}, 
but also concept-level TBox answers as squares: \textit{AI Researchers},
\textit{The Academia}, and \textit{The Industry}. }
\vspace{-0.5cm}
  \label{fig:example}
\end{figure}

Along with the rapid development of high-quality large-scale knowledge
infrastructures \cite{tanon2020yago, auer2007dbpedia}, researchers are
increasingly interested in exploiting knowledge bases for real-world
applications, such as knowledge graph completion
\cite{bordes2013translating, tang2022positive} and entity alignment
\cite{trisedya2019entity}.  However, to take advantage of knowledge
bases, a fundamental yet challenging task still remains unsolved,
i.e., 
neural logical reasoning (NLR), which attempts to answer complex
structured queries that include logical operations and multi-hop
projections given the facts in knowledge bases with distributed representations
\cite{hamilton2018embedding}. Recently, efforts
\cite{hamilton2018embedding, ren2019query2box, ren2020beta} have been
made to develop NLR systems by designing strategies to
learn geometric or uncertainty-aware distributed query
representations, and proposing mechanisms to deal with various logical
operations on these distributed representations.

However, existing neural logical reasoners cannot fully fulfill our
needs. In many real-world scenarios, we not only expect
entity-level answers, but also seek for more descriptive concept-level
answers, where each of the concepts is a description of
a set of entities. For example, as shown in Figure \ref{fig:example},
the query asks
``who will be interested in techniques that G. Hinton is investigating
and Google is using?'', users expect entity-level answers like
\textit{Meta}, \textit{Amazon}, \textit{MIT}, and \textit{Y. LeCun},
as well as concept-level answers such as \textit{AI Researchers},
\textit{The Academia}, and \textit{The Industry}.  In this example,
the conceptual answer \textit{The Academia} refers to a summary of a
set consisting of \textit{Y. LeCun} and \textit{MIT}, and it is
intuitively desirable for users and worth exploring. In biomedical
applications, people may want to find the causes for a set of symptoms
and expect both entity-level answers (such as \textit{SARS-CoV-2} causing
\textit{Fever}) as well as concept-level answers (such as  \textit{Viral
  infections} causing \textit{Fever}).  In this case, the answer
constitutes a descriptive concept-level answer (e.g., \textit{Viral
  infections}) that is a summary of a set of entity-level
answers. Downstream tasks like online chatbots \cite{liu2018knowledge}
and conversational recommender systems \cite{zhou2020improving} also
need to retrieve rich and comprehensive answers to provide better
services.  Thus, providing both entity-level and concept-level answers
can highly improve their capability of generating
more informative responses to users and enriching the semantic
information in answers for downstream tasks.

From the perspective of ontologies, i.e., description logic (DL) based knowledge bases, the ability of providing
both \textbf{concept}-level and \textbf{entity}-level answers corresponds
to the capability of TA-NLR: \underline{n}eural
\underline{l}ogical \underline{r}easoning across \textbf{\underline{T}Box}, i.e., terminology axioms between two concepts, and
\textbf{\underline{A}Box}, i.e., assertion axioms
between two entities or between a concept and an entity \cite{baader2003description}.
Previous NLR systems only support reasoning over ABox, while more general
TA-NLR systems additionally support reasoning over the whole ontology which is highly useful in applied ontologies \cite{elsenbroich2006case}.  
Therefore, the TA-NLR problem is more general than the regular NLR
problem in terms of providing not only entity-level ABox answers, but also concept-level TBox answers.  
Note that such descriptive concepts are
higher-level abstractions of the set of entities and are more
informative than the set in some cases \cite{cook2009dictionary}.

Therefore, from the perspective of users and downstream tasks, TA-NLR is helpful by jointly providing the more informative \textbf{concept}-level and \textbf{entity}-level answers.
From the perspective of logic theory, TA-NLR is useful in ontological applications by jointly reasoning over \textbf{TBox} and \textbf{ABox}.  
However, existing methods \cite{hamilton2018embedding, ren2019query2box,
  ren2020beta} can hardly reach TA-NLR for the following reasons.  On
the one hand, concepts are excluded from the NLR systems. That is to
say, previous NLR systems perform reasoning upon regular
knowledge graphs where only entities and relations exist, which correspond to subsets of ABoxes.  On the other hand,
mechanisms for exploiting concepts have not been well established. Specifically, previous solutions only measure query-entity similarity for NLR, without considering concept representations and operators involving concepts, such as query--concept similarity.

Along this line, we propose an original solution named  \textbf{TAR} for
TA-NLR. \textbf{TAR} is a shortname which stands for a \textbf{T}Box and \textbf{A}Box neural \textbf{r}easoner.
The key challenges for addressing TA-NLR are the \textbf{incorporation of concepts, representation of concepts, and operator on concepts}.
First, we observe that terminological axioms include taxonomic hierarchies of concepts,
concept definitions, and concept subsumption relations
\cite{gruber1993translation}. To \textbf{incorporate concepts} into the TA-NLR system, we thus introduce some terminological axioms into the system to provide sources of
concepts.
Second, we find that
fuzzy sets \cite{klir1995fuzzy}, i.e., sets whose elements have
degrees of membership, can naturally bridge entities with concepts, i.e., vague sets of entities. 
Therefore, we \textbf{represent concepts} as fuzzy sets in TAR.  Meanwhile, properly
representing queries is the prerequisite of effectively
operating on concepts. We find that fuzzy sets can also bridge entities with queries, i.e., vague sets of entity-level answers. The
theoretically-supported, vague, and unparameterized fuzzy set
operations enable us to resolve logical operations within queries. Thus, the adoption of fuzzy sets is an ideal solution for concepts and queries representation in TA-NLR. 
Then, \textbf{operators} involving concepts can also be designed based on fuzzy sets, including
query-concept operators for abduction, entity-concept operators for
instantiation, and concept-concept operator for subsumption. 
Attributed to the well designed operators, a joint TBox and Abox neural reasoner
can be achieved. 

We summarize the main contribution of this work as follows: (1) To the
best of our knowledge, we are the first to focus on the TA-NLR problem
that aims at providing both entity-level ABox answers and
concept-level TBox answers, which better satisfies the need
of users, downstream tasks, and ontological applications;\\
(2) We propose an original solution TAR that properly incorporates,
represents, and operates on concepts. We incorporate terminological
axioms to provide sources of concepts and employ fuzzy sets as the
representations of concepts and queries. Logical operations are
supported by the well-established fuzzy set theory and
operators involving concepts are rationally designed upon fuzzy sets;\\
(3) We conduct extensive experiments and demonstrate the effectiveness
of TAR for TA-NLR. We publish in public two pre-processed benchmark
datasets for TA-NLR and the implementation of
TAR\footnote{https://anonymous.4open.science/r/TAR-5D7D} to foster
further research.


\section{Related Work}
\subsection{Neural Logical Reasoning}
Given the vital role of neural logical reasoning (NLR) in knowledge
discovery and artificial intelligence, great efforts have been made to
develop NLR systems recently.  GQE \cite{hamilton2018embedding} is the
pioneering work in this field, the authors formulate the NLR problem
and propose to simply use points in the embeddings space to represent
logical queries. Q2B \cite{ren2019query2box} claimed that the
representation of each query in the embedding space should be a
geometric region instead of a single point because each query is
equivalent to a set of entity-level answers in the embedding
space. Therefore, they use hyper-rectangles that can include multiple
points in the embedding space to represent queries. HypE
\cite{choudhary2021self}, ConE \cite{zhang2021cone}, and BetaE
\cite{ren2020beta} extended Q2B by using more sophisticated geometric
shapes or Beta distributions for query representation.
However, these reasoners could only give extensional entity-level
answers, while we focus on the more general TA-NLR problem that aims
at additionally providing descriptive concepts. We bring neural logical reasoners to the
stage of reasoning across TBox and ABox.


\subsection{Fuzzy Logic for NLR}
Besides representing logical queries as points, geometric regions, or
distributions, more recent methods explore fuzzy logic
\cite{klir1995fuzzy} for NLR.  CQD \cite{arakelyan2020complex} used
$t$-norm and $t$-conorms from the fuzzy logic theory to achieve high
performance on zero-shot settings. More specifically,   
mechanisms are proposed for the inference stage on various types of queries, while
only training the simple neural link predictor on triples (\textit{1p}
queries in Figure \ref{fig:types}). FuzzQE \cite{chen2022fuzzy}, GNN-QE \cite{zhu2022neural}, and LogicE \cite{luus2021logic}
directly represent entities and queries using embeddings with
specially designed restrictions and interpreted them as fuzzy sets for
NLR. However, these studies still focus on the regular NLR problem, while we
are solving a more general problem that additionally gives concept-level TBox answers. Furthermore, we explicitly include \emph{concepts} and
represent them as fuzzy sets, whereas they   represent only  \emph{queries} as fuzzy sets. Moreover, they either just use fuzzy logic at the entity level, or use fuzzy sets with arbitrary numbers of elements as the
tunable embedding dimension without reasonable interpretations.  We
interpret queries as fuzzy sets where each element represents the
probability of an entity being an answer,    aligning with the
definition and the essence of fuzzy sets \cite{klir1995fuzzy}, i.e.,
sets where each element has a degree of membership. This allows us to
fully exploit fuzzy logic and provides a theoretical foundation in
fuzzy set theory.

\subsection{Ontology Representation Learning}
\begin{figure}[t!]
  \centering \includegraphics[width=0.9\linewidth]{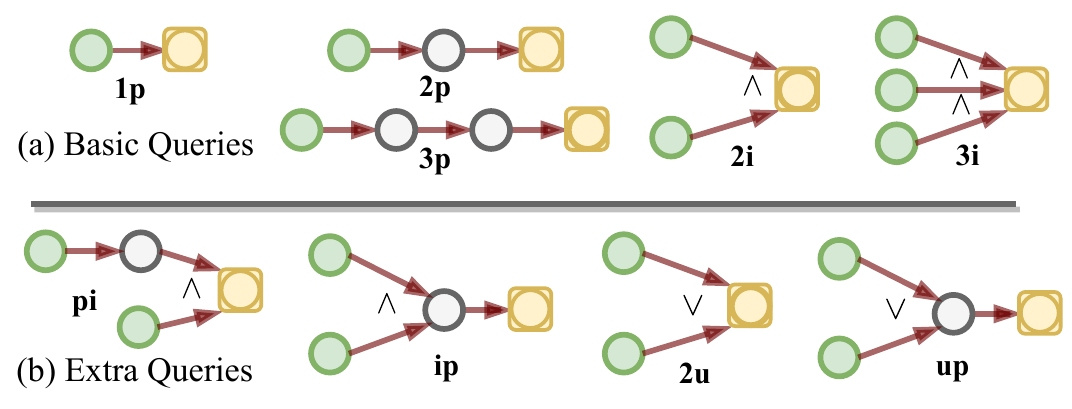}
  \caption{The considered types of queries represented with their
    graphical structures. $\wedge$ and $\vee$ represent the intersection and union logical operations, respectively. Squares denote concepts
    and circles represent entities.}
  \label{fig:types}
\end{figure}

Several methods have been developped to exploit ontologies from
  the perspective of distributed representation learning  \cite{kulmanov2021semantic}.  ELEm \cite{kulmanov2019embeddings} and EmEL
\cite{mondala2021emel} learn geometric embeddings for concepts in
ontologies. The key idea of learning geometric embeddings is that the
embedding function projects the symbols used to formalize \el axioms
into an interpretation $\mathcal{I}$ of these symbols such that
$\mathcal{I}$ is a model of the \el ontology.  Other approaches
\cite{smaili2019opa2vec,chen2021owl2vec} rely
on regular graph embeddings or word embeddings and apply them to
ontology axioms. Another line of research \cite{hao2019universal,
  hao2020bio} focuses on jointly embedding entities and relations in
regular knowledge graphs, as well as concepts and roles (relations) in
ontological axioms.  Our work is related to ontology representation
learning in that we incorporate some description logic based
ontological axioms in Section \ref{GeneralMLR} to provide sources of
concepts, and we exploit concepts with distributed
representation learning in our proposed TAR for TA-NLR.
Methods for representation learning with ontologies have
previously only been used to answer link prediction tasks such as
predicting protein--protein interactions or performing knowledge graph
completion, which can be viewed as answering {\em 1p} queries in
Figure \ref{fig:types} whereas we also focus on more complex queries
as well as providing concept-level TBox answers.

\section{Methodology}
Incorporating, representing, and operating on concepts
are the key components for a neural
logical reasoner across ABox and TBox.  In this section, we first formulate the TA-NLR
problem along with the process of incorporating concepts into the
reasoning system. Then we propose an original solution \textbf{TAR}
for TA-NLR by designing concept representations and 
operators involving concepts. We introduce optimization and inference procedures in the
end.

\begin{figure}[t!]
  \centering \includegraphics[width=0.5\linewidth]{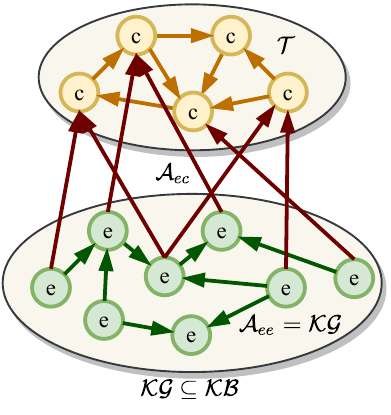}
  \vspace{-0.3cm}
  \caption{Illustration of the description logic based ontological knowledge base $\mathcal{KB} = (\mathcal{T}, \{\mathcal{A}_{ee}, \mathcal{A}_{ec}\})$.}
  \label{fig:kb}
\end{figure}

\subsection{Incorporating Concepts}

\subsubsection{\textbf{Regular NLR}}
\label{MLR}
The regular NLR problem is defined on knowledge graphs. A knowledge
graph is formulated as
$\mathcal{KG} = \{ \langle h, r, t \rangle \} \subseteq \mathcal{E}
\times \mathcal{R} \times \mathcal{E}$, where $h$, $r$, $t$ denote the
head entity, relation, and tail entity in triple
$\langle h, r, t \rangle$, respectively, $\mathcal{E}$ and
$\mathcal{R}$ refer to the entity set and the relation set in
$\mathcal{KG}$.

In the context of NLR, as shown in Figure \ref{fig:types}.(a), each
triple $\langle h, r, t \rangle$ is regarded as a positive sample of
the {\em 1p} query $\exists ?: [h \xrightarrow{r}](?)$ with an answer
$t$ that satisfies $[h \xrightarrow{r}] (t)$, where $h$ is the anchor
entity and $\xrightarrow{r}$ is the projection operation with relation
$r$. Furthermore, the regular NLR problem may also
  address the intersection, union, and negation
operations $\wedge$, $\vee$, and $\neg$ within queries. Thus, infinite
types of queries can be found with the combinations of these logical
operations. We consider the representative types of queries, which are
listed and demonstrated with their graphical structures in Figure
\ref{fig:types}. For example, queries of type {\em pi} in Figure
\ref{fig:types}.(c) are to ask
$\exists ?: [(h_1 \xrightarrow{r_1} \xrightarrow{r_2}) \wedge (h_2
\xrightarrow{r_3})](?)$.

Regular neural logical reasoners seek to provide
entity-level answers for each query. In
particular, the answers are a set of entities that satisfy the query. 
We predict the possibility of each candidate entity
$e \in \mathcal{E}$ satisfying a query $\exists ?: [q](?)$. We then
rank the $|\mathcal{E}|$ possibilities and select the top-$k$ entities
in $\mathcal{E}$ as the set of answers.  Since all the candidate
answers are entities, we can only retrieve entity-level
answers from the regular NLR systems.

\subsubsection{\textbf{NLR across TBox and ABox}}
\label{GeneralMLR}
A joint TBox and ABox neural logical reasoner is upon a description logic based knowledge base $\mathcal{KB}$, i.e., ontology, which
is an ordered pair ($\mathcal{T}$, $\mathcal{A}$) for TBox
$\mathcal{T}$ and ABox $\mathcal{A}$, where $\mathcal{T}$ is a finite
set of terminological axioms and $\mathcal{A}$ is a finite set of
assertion axioms. Specifically, terminological axioms within a TBox
$\mathcal{T}$ are of the form $c_1 \sqsubseteq c_2$ where the symbol
$\sqsubseteq$ denotes subsumption ($subClassOf$). In
  general, $c_1$ and $c_2$ can be concept descriptions that consist
  of concept names, quantifiers and roles (relations), and logical
  operators; we limit TAR to axioms where $c_1$ and $c_2$ are concept
  names that will not involve roles or logical operators
  \cite{baader2003appendix}. In the followings, we do not distinguish
  between a concept name and a concept description unless there are
  special needs. Then, a TBox is:
\begin{equation}
  \mathcal{T} \subseteq  \{  c_i \sqsubseteq c_j | c_i,c_j \in \mathcal{C} \}
  \label{tbox}
\end{equation}
where $\mathcal{C}$ denotes the set of concept names in
$\mathcal{KB}$.  $\mathcal{T}$ accounts for the source of concepts and
the pairwise concept subsumption information in the TA-NLR system.
Assertion axioms in $\mathcal{A}$ consist of two parts.  One part
is the role assertion that is expressed as:
\begin{equation}
  \mathcal{A}_{ee} \subseteq \{ \langle e_1, r, e_2 \rangle | e_1,e_2
  \in \mathcal{E}, r \in \mathcal{R} \}
  \label{aboxee}
\end{equation}
where $e_1, e_2 \in \mathcal{E}$ denote entities, $\mathcal{E}$
denotes the entity set in $\mathcal{KB}$, $r \in \mathcal{R}$ denotes
the role assertion between $e_1$ and $e_2$, and $\mathcal{R}$ is the
the role set of $\mathcal{A}_{ee}$. $\mathcal{A}_{ee}$ accounts for
the triple-wise relational information about entities and roles in
TA-NLR.  The other part within $\mathcal{A}$ is the concept
instantiation between an entity $e \in \mathcal{E}$ and a concept
$c \in \mathcal{C}$:
\begin{equation}
  \mathcal{A}_{ec} = \{ e \triangleleft c  \} \subseteq \mathcal{E} \times \mathcal{C},
  \label{aboxec}
\end{equation}
where $e \triangleleft c$ represents $e$ is an instance of
$c$. $\mathcal{A}_{ec}$ serves as the bridge between $\mathcal{T}$ and $\mathcal{A}_{ee}$
by providing pairwise links between entities and concepts.

Since we incorporate concepts in the TA-NLR systems, we are able to
ask questions about concepts. In particular, for a query
$\exists ?: [q](?)$ of arbitrary type discussed in Section \ref{MLR},
we not only provide a set of entities
of $\{a_e\}$ as the entity-level ABox answers, but also infer an explanation for each query result
by summarizing entity-level answers with descriptive concepts,
yielding another set of concept-level answers $\{a_c\}$ as the 
concept-level TBox answers. More specifically, as shown in Figure
\ref{fig:types}, the answers are no longer restricted to be
$e \in \mathcal{E}$ (denoted by circles), they can also be
$c \in \mathcal{C}$ (denoted by squares).  To achieve this goal,
we predict the possibility of each candidate entity
$e \in \mathcal{E}$ as well as the possibility of each candidate
concept $c \in \mathcal{C}$ satisfying a query $\exists ?: [q](?)$.
We then rank $|\mathcal{E}|$ predicted scores of candidate entities
and $|\mathcal{C}|$ predicted scores of candidate concepts. We select
and combine the top-$k$ results from each set of candidates as the
final answers of $q$ with entity-level and
concept-level answers $\{a\} = \{a_e\} \bigcup \{a_c\}$.

Note that the regular NLR problem is a sub-problem of the TA-NLR
problem.  First, regular NLR systems can only provide a
subset of the answers provided by TA-NLR systems, i.e., $\{a_e\} \subseteq \{a\}$.  Also, the entire $\mathcal{KG}$ in the context of regular NLR is
equivalent to $\mathcal{A}_{ee}$ in the case of the TA-NLR problem,
which is a subset of the ontology, i.e.,
$\mathcal{KG} \subseteq \mathcal{KB}$, leaving $\mathcal{T}$ and
$\mathcal{A}_{ec}$ with conceptual information in the ontologies not
explored.  Therefore, the problem we investigate is more general in
terms of providing more answers and reasoning over more complex
knowledge bases.

\subsection{Representing Concepts and Queries}
In this subsection, we first introduce how to represent concepts as
fuzzy sets in our proposed  TAR for
TA-NLR. Then we represent queries
as fuzzy sets as well to prepare for the later operations that involve
concepts and queries.

\begin{figure}[t!]
  \centering \includegraphics[width=0.6\linewidth]{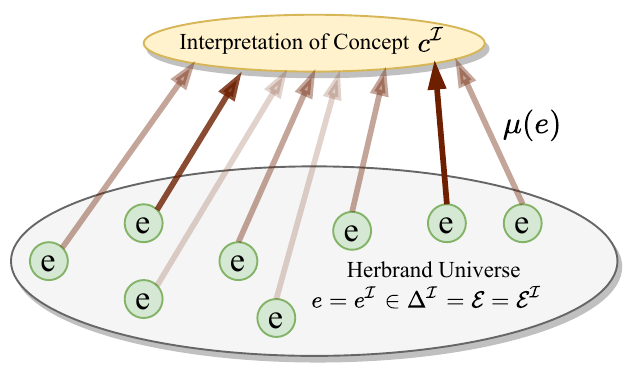}
  \caption{Illustration of the interpretation of a concept $c^\mathcal{I}$.}
  \label{fig:concept}
\end{figure}

\subsubsection{\textbf{Representing Concepts}}
\label{representconcept}
We are motivated to represent concepts as fuzzy sets by the
relationship between concepts and entities. We gain insights on such
relationship from the basic definition of semantics in description
logics \cite{baader2008description}:
\begin{definition}
  A terminological interpretation
  $\mathcal{I}=\left(\Delta^{\mathcal{I}}, \cdot^{\mathcal{I}}\right)$
  over a signature
  $\left(\mathcal{C}, \mathcal{E}, \mathcal{R}\right)$ consists of:
\begin{itemize}
\item a non-empty set $\Delta^{\mathcal{I}}$ called the domain
\item an interpretation function ${\cdot}^{\mathcal{I}}$ that maps:
  \begin{itemize}
  \item every entity $e \in \mathcal{E}$ to an element
    $e^{\mathcal{I}} \in \Delta^{\mathcal{I}}$
  \item every concept $c \in \mathcal{C}$ to a subset of
    $\Delta^{\mathcal{I}}$
  \item every role (relation) $r \in \mathcal{R}$ to a subset of
    $\Delta^{\mathcal{I}} \times \Delta^{\mathcal{I}}$
  \end{itemize}
\end{itemize}
\label{definition}
\end{definition}

As we use a function-free language \cite{baader2003appendix}, we set $\Delta^{\mathcal{I}}$ to be the
Herbrand universe \cite{lee1972fuzzy} of our knowledge base, i.e.,
  $\Delta^{\mathcal{I}} = \mathcal{E}$. Therefore, according to Definition \ref{definition}, 
  the interpretation of concept $c^\mathcal{I}$ is a subset of $\mathcal{E}$, which is
  finite. 
On the other hand, fuzzy sets \cite{klir1995fuzzy} over
  the Herbrand Universe are finite
  sets whose
elements have degrees of membership:
\begin{equation}
  FS = \{ \mu(x_1), \mu(x_2), \cdots, \mu(x_{|FS|}) \},
\end{equation}
where $\mu(\cdot)$ is the membership function that measures the degree of membership of each element.
Therefore, we further interpret all concepts as fuzzy sets over the finite domain $\Delta^{\mathcal{I}} = \mathcal{E} = \{e_1, e_2, \cdots, e_{|\mathcal{E}|}\}$ as the elements of fuzzy sets $\{x_1, x_2, \cdots, x_{|FS|}\}$. Thus, we have:
\begin{equation}
  c^\mathcal{I} = \{ \mu(e_1), \mu(e_2), \cdots, \mu(e_{|\mathcal{E}|}) \}.
  \label{eqn:fsconcept_interpretation}
\end{equation}
As the Herbrand universe for our language is always finite, the
interpretation of concept $c^\mathcal{I}$ is fully determined by the
fuzzy membership function $\mu(\cdot)$ that assigns a degree of
membership to each entity
$e = e^\mathcal{I} \in \Delta^{\mathcal{I}} = \mathcal{E} =
\mathcal{E}^\mathcal{I}$ for
$c^\mathcal{I} \in \mathcal{C}^\mathcal{I}$, where
$\mathcal{E}^\mathcal{I}$ and $\mathcal{C}^\mathcal{I}$ are the
interpretation of the entity set and the concept set.

To obtain the degree of membership of 
entity $e_i$ in $c^\mathcal{I}$, i.e., $\mu(e_i)$, we first randomly
initialize the embedding matrix of concepts and entities as
$\mathbf{E}_c \in \mathbb{R}^{|\mathcal{C}| \times d}$ and
$\mathbf{E}_e \in \mathbb{R}^{|\mathcal{E}| \times d}$ with Xavier
uniform initialization \cite{glorot2010understanding}, where $d$ is
the embedding dimension. Then we obtain the embedding of each concept
$\mathbf{c} \in \mathbb{R}^d$ by looking up the rows of $\mathbf{E}_c$. The embedding then serves as
the generator of the fuzzy set representation of each concept
$FS_c$. Thus, we compute the similarities between each concept $c$ and
every entity
  in our universe
$e \in \mathcal{E} = \Delta^{\mathcal{I}}$ as the degrees of
membership of each entity in the fuzzy set:
\begin{equation}
  FS_c = \{ \sigma( \mathbf{c} \otimes \mathbf{E}_e^T) \} = c^\mathcal{I},
  \label{eqn:fsconcept}
\end{equation}
where symbol $\otimes$ denotes matrix multiplication and ${\cdot}^T$
represents the matrix transposition. The measured similarities are
then normalized to $(0, 1)$ using the bit-wise sigmoid function
$\sigma(\cdot)$. Here, the set-wise operation to obtain $FS_c$
consists of $|\mathcal{E}|$ pair-wise operations on the
entity--concept pairs; we use the same operator for
\textit{Instantiation}, which we will explain in Section
\ref{instantiation}.

\subsubsection{\textbf{Representing Queries}}
Properly representing queries is the prerequisite of operating on
concepts. Fuzzy sets are particularly suitable to represent not only
concepts, but also queries,
because interpretations of queries are essentially interpretations of
concepts. More accurately, queries correspond to concept descriptions
that may include concept names, roles (relations), quantifiers, and
logical operations.  We can use the same formalism designed for
representing concepts to represent entities, i.e., as a special type
of fuzzy set \cite{rihoux2009crisp} that assigns the membership
function $\mu(\cdot)$ to $1$ for one exact  entity and to $0$ to all
others.  Consequently, we can interpret entities as concepts.  As
explained in section \ref{MLR}, queries may consist of entities, relations,
and logical operations.  Therefore, queries are interpreted as concept
descriptions and we regard \textbf{entities} within queries as
singleton concepts.  Thus, we can use the same description logic
semantics \cite{baader2017introduction} to interpret a query $q$ and a
concept $c$ in Definition \ref{definition}: an interpretation function
${\cdot}^{\mathcal{I}}$ maps every query $q$ to a subset of
$\Delta^{\mathcal{I}}$. As the Herbrand universe
$\Delta^{\mathcal{I}} = \mathcal{E}$ is finite, the interpretation of
query $q^\mathcal{I}$ is fully determined by the fuzzy membership
function
\begin{equation}
  q^\mathcal{I} = \{ \mu(e_1), \mu(e_2), \cdots, \mu(e_{|\mathcal{E}|}) \}.
  \label{eqn:fsquery_interpretation}
\end{equation}
Besides, representing queries as fuzzy sets has other advantages.
Firstly, fuzzy logic theory
\cite{klir1995fuzzy} well-equips us to interpret logical operations
within queries as the vague and unparameterized fuzzy set
operations. The preservation of vagueness is important in that TA-NLR requires
uncertainty, rather than deductive reasoning that guarantees the
correctness. Unparameterized operations are desirable because
they require fewer data during training and are often
  more interpretable.  Secondly, since concepts are
already represented as fuzzy sets, it would be more convenient for us
to employ the same form of representation and retain only one form of
representation within the TA-NLR system.  We explain how to represent
queries as fuzzy sets in detail as the followings.

\paragraph{\textbf{Representing Atomic Queries}}
Each multi-hop logical query consists of one or more Atomic Queries
(AQ), where an AQ is defined as a query that only contains
projection(s) $\ \xrightarrow{r} \ $ from an anchor entity without
logical operations such as intersection $\wedge$, union $\vee$, and
negation $\neg$. Therefore, the first step to represent queries is to
represent AQs.  We obtain the embeddings of each entity
$\mathbf{e} \in \mathbb{R}^d$ and the $i^{th}$ relation
$\mathbf{r} \in \mathbb{R}^d$ by looking up the rows of the randomly
initialized entity embedding matrices
$\mathbf{E}_e \in \mathbb{R}^{|\mathcal{E}| \times d}$ and
$\mathbf{E}_r \in \mathbb{R}^{|\mathcal{R}| \times d}$ with Xavier
uniform initialization \cite{glorot2010understanding}. Then, the
generator for fuzzy set representation $FS_{aq}$ of a valid AQ
$[e \xrightarrow{r_1} \cdots \xrightarrow{r_i}](?)$ is
$(\mathbf{e} + \mathbf{r_{1}} + \cdots + \mathbf{r_{i}})$. Thus, we
obtain the fuzzy set corresponding to the query $aq$ as:
\begin{equation}
  FS_{aq} = \{ \sigma((\mathbf{e} + \mathbf{r_{1}} + \cdots + \mathbf{r_{i}}) \otimes \mathbf{E}_e^T) \} = aq^\mathcal{I}.
  \label{eqn:fsaq}
\end{equation}
Similar to the process of obtaining fuzzy set representations of
concepts, Eq.(\ref{eqn:fsaq}) is to aquire the degrees of membership
of every candidate $e \in \mathcal{E}$ being an answer
  to a given AQ by computing their normalized similarities.

\begin{figure}[t]
  \centering \includegraphics[width=0.8\linewidth]{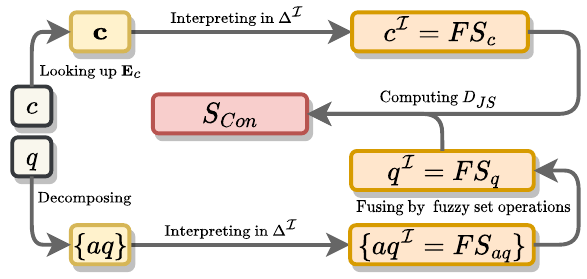}
  \vspace{-0.1cm}
  \caption{Illustration of the process of representing concepts and queries to provide concept-level TBox answers.}
  \label{fig:abduction}
  \vspace{-0.1cm}
\end{figure}

\paragraph{\textbf{Fusing Atomic Queries}}
AQs are fused by logical operations to form multi-hop logical
queries. Since AQs are already represented in fuzzy sets and we are
equipped with the theoretically supported fuzzy set operations, we
interpret logical operations as fuzzy set operations over concepts to
fuse AQs into the final query representations.

For two fuzzy sets in domain
  $\Delta^\mathcal{I} = \mathcal{E}$:
  $FS_1 = \{ \mu_1(e_1), \cdots, \mu_1(e_{|\mathcal{E}|}) \}$ and
  $FS_2 = \{ \mu_2(e_1), \cdots, \mu_2(e_{|\mathcal{E}|}) \}$, we have
  the \textbf{intersection} $\wedge$ over the two fuzzy sets as:
\begin{multline}
   FS_\wedge =  \{ \mu_\wedge(e_1), \cdots, \mu_\wedge(e_{|\mathcal{E}|}) \} \\ = FS_1 \wedge FS_2 = \{ \forall e \in \mathcal{E}: \mu_{\wedge}(e)=\top \left(\mu_{1}(e), \mu_{2}(e)\right) \},
  \label{inter}
\end{multline}
the \textbf{union} $\vee$ over the two fuzzy sets as:
\begin{multline}
    FS_\vee = \{ \mu_\vee(e_1), \cdots, \mu_\vee(e_{|\mathcal{E}|}) \}
    \\ =  FS_1 \vee FS_2 = \{ \forall e \in \mathcal{E}: \mu_{\vee}(e)= \ \perp \left(\mu_{1}(e), \mu_{2}(e)\right) \},
  \label{union}
\end{multline}
and we have the \textbf{negation} $\neg$ over $FS$ as:
\begin{multline}
\small
  FS_\neg = \{ \mu_\neg(e_1), \cdots, \mu_\neg(e_{|\mathcal{E}|}) \}  =  \{ \forall e \in \mathcal{E}: \mu_{\neg}(e)=1-\mu(e) \},
  \label{neg}
\end{multline}
where a $t$-norm $\top:[0,1] \times[0,1] \mapsto[0,1]$ is a
generalisation of conjunction in logic
\cite{klement2004triangular}. Some examples of $t$-norms include the
Gödel $t$-norm $\top_{\min }(x, y)=\min \{x, y\}$, the product $t$
norm $\top_{\text {prod }}(x, y)=x \cdot y$, and the Łukasiewicz
$t$-norm $\top_{\text {Łuk }}(x, y)=\max \{0, x+y-1\}$
\cite{van2020analyzing}. Analogously, a $t$-conorm
$\perp:[0,1] \times[0,1] \mapsto[0,1]$ is dual to $t$-norm and
generalizes logical disjunction -- given a $t$-norm $\top$, the
complementary $t$-conorm is defined by $\perp(x, y)=1-\top(1-x, 1-y)$
\cite{arakelyan2020complex}. The choice of the $t$-norm is a
hyperparameter of TAR.

Thus, each query can be decomposed into AQs and represented as a fuzzy
set with Eq.(\ref{eqn:fsaq}), and then fuzzy set representations of
AQs are fused by the fuzzy set operations in Eq.(\ref{inter}),
(\ref{union}), and (\ref{neg}) to obtain the final representation of
the query. Note that fuzzy set operations hold the property of
\textit{closure}, which means the input and output of these operations
remain fuzzy sets. Thus, the final representation of each query is
also a fuzzy set $FS_q$. 

\subsection{Operating on Concepts}
\label{operator}
In previous sections, we manage to prepare for designing operators
involving concepts by representing concepts and queries in fuzzy
sets. Here, we design operators involving concepts for concept retrieval,
entity retrieval, subsumption, and instantiation.

\subsubsection{\textbf{Concept Retrieval}}
\label{QA}
Concept retrieval is to provide concept-level TBox answers, i.e., $\{ a_c \}$ as discussed in
Section \ref{GeneralMLR}.  We measure the possibility of each
$c \in \mathcal{C}$ being an intensional concept-level answer of a
given query upon fuzzy set representations. More specifically, we
measure the similarity between $FS_c$ and $FS_q$ based on the
Jensen-Shannon divergence $D_{JS}$ \cite{endres2003new}, which is a
symmetrized and smoothed version of the Kullback-Leibler divergence
$D_{KL}$.  The similarity function $S_{Con}$ is defined by:
\begin{equation}
  S_{Con} = - D_{JS}(P \| Q)= - \frac{1}{2} D_{KL}(P \| M)+\frac{1}{2} D_{KL}(Q \| M)
\label{sabd1}
\end{equation}
where $M=\frac{1}{2}(P+Q)$, $P$ and $Q$ represent the normalized fuzzy
set representations of the considered query and concept descriptions,
which are given by:
\begin{equation}
  P = \frac{FS_c}{\max \left(\|FS_c\|_{p}, \epsilon\right)},
  Q = \frac{FS_q}{\max \left(\|FS_q\|_{p}, \epsilon\right)},
\label{sabd2}
\end{equation}
where $\epsilon$ is a small value to avoid division by zero and $p$ is
the exponent value in the norm formulation $\|\cdot\|_p$. $S_{Con}$
is then used for model training and concept-level inference in Section
\ref{train}.

\subsubsection{\textbf{Entity Retrieval}} \label{sec:e-retrieval}
Entity retrieval aims to provide entity-level ABox answers;
for this purpose, only query--entity similarities $S_{Ent}$ need to be
measured without the necessity of designing new mechanisms. Therefore,
we follow the pioneering work \cite{hamilton2018embedding} on NLR to represent each query as an embedding $\mathbf{q} = f(q; \Omega)$
and measure query--entity similarity $S_{Ent}$ as: \vspace{-0.05cm}
\begin{equation}
    S_{Ent} = \gamma  - \|\mathbf{q} - \mathbf{e}\|_1
\label{sind}  \vspace{-0.05cm}
\end{equation}
where $\gamma$ is the margin, $f(\cdot)$ denotes the function to obtain query embedding $\mathbf{q}$, and $\Omega$ denotes the parameters of $f(\cdot)$. We explain $f(\cdot)$ in detail in supplementary materials\footnote{https://anonymous.4open.science/r/TAR-5D7D/Appendix.pdf}.
\begin{table*}[htbp]
\small
    \renewcommand\arraystretch{0.9}
  	\renewcommand\tabcolsep{2.6pt}
  \centering
  \caption{\textbf{Con}, \textbf{Ent}, \textbf{Sub}, \textbf{Ins}, and \textbf{NLR} correspond to statistics of the instances for Concept Retrieval, Entity Retrieval, Subsumption, Induction, and baseline NLR methods. For \textit{other} query types in
    Figure \ref{fig:types}, the statistics are the same for each of
    them.}
     \vspace{-0.25cm}
    \begin{tabular}{c||ccc||c||cc|cc|c|c||cc}
    \toprule
          & $|\mathcal{E}|=|FS|$    & $|\mathcal{C}|$      & $|\mathcal{R}|$      & Partition & \textbf{Con}-1p & \textbf{Con}-other & \textbf{Ent}-1p & \textbf{Ent}-other & \textbf{Sub} & \textbf{Ins} & \textbf{NLR}-1p & \textbf{NLR}-other \\
    \midrule
    \multirow{2}[1]{*}{YAGO4} & \multirow{2}[1]{*}{32,465} & \multirow{2}[1]{*}{8,382} & \multirow{2}[1]{*}{75} & \#Train & 189,338 & 10,000 & 101,417 & 10,000 & 16,644 & 83,291 & 184,708 & 10,000 \\
          &       &       &       & \#Valid/\#Test & 1,000/1,000 & 1,000/1,000 & 1,000/1,000 & 1,000/1,000 & -     & -     & 1,000/1,000 & 1,000/1,000 \\
    \midrule
    \multirow{2}[2]{*}{DBpedia} & \multirow{2}[2]{*}{28,824} & \multirow{2}[2]{*}{981} & \multirow{2}[2]{*}{327} & \#Train & 473,924 & 10,000 & 136,821 & 10,000 & 2,582 & 225,436 & 362,257 & 10,000 \\
          &       &       &       & \#Valid/\#Test & 1,000/1,000 & 1,000/1,000 & 1,000/1,000 & 1,000/1,000 & -     & -     & 1,000/1,000 & 1,000/1,000 \\
    \bottomrule
    \end{tabular}%
  \label{statistic}%
\end{table*}%

\subsubsection{\textbf{Subsumption}}
\label{subsumption}
As defined by Eq.(\ref{tbox}), $\mathcal{T}$ supplies for
relational information among concepts with the form of concept
subsumptions. Although concepts are represented in fuzzy sets and we
already designed mechanism to measure the similarity between two fuzzy
sets, we can not directly apply the method in Section \ref{QA} for
concept subsumptions. It is because we need to measure the degree of
inclusion of one concept to another instead of the similarities
between them. The degree of inclusion is asymmetrical and more complex than the
similarity measurement. Therefore, we employ a neural network
$h(\cdot)$ to model the degree of inclusion:
\begin{equation}\vspace{-0.05cm}
  S_{Sub}=h\left(\mathbf{c_1} \oplus \mathbf{c_2}  ; \theta\right) 
 \label{ssub} \vspace{-0.05cm}
\end{equation}
where symbol $\oplus$ denotes matrix concatenation over the last
dimension, and $\theta$ denotes the parameters of $h(\cdot)$. In this
paper, $h(\cdot)$ is a two-layer feed-forward network with $Relu$
activation. Note that we directly use the embeddings of concepts without interpreting concept in the Herbrand universe of entities $\Delta^\mathcal{I} = \mathcal{E}$ because neither concept-entity relationships
need to be modeled nor logical operations need to be resolved.

\subsubsection{\textbf{Instantiation}}
\label{instantiation}
As defined by Eq.(\ref{aboxec}), $\mathcal{A}_{ec}$ bridges
$\mathcal{T}$ and $\mathcal{A}_{cc}$ by providing links between
entities and concepts. Such links instantiate concept with its
describing entities and thus offer relational information with the
form of concept instantiation. Recall that in Section
\ref{representconcept}, we obtain the fuzzy set representation of
concepts by computing the similarities between the given $c$ and every
candidate $e \in \mathcal{E}$ with Eq.(\ref{eqn:fsconcept}). In the
case of concept instatiation, the set-wise computation
Eq.(\ref{eqn:fsconcept}) is degraded to pair-wise similarity
measurement for each concept-entity pair:
\begin{equation}
  S_{Ins} = \sigma(\mathbf{c} \otimes \mathbf{e}^T)
  \label{sins}
\end{equation}
where $\mathbf{c} \in \mathbb{R}^d$ and $\mathbf{e} \in \mathbb{R}^d$
are the categorical embeddings of concept $c$ and entity $e$,
respectively.

\subsection{\textbf{Optimization}}
\begin{figure}[t]
  \centering \includegraphics[width=0.7\linewidth]{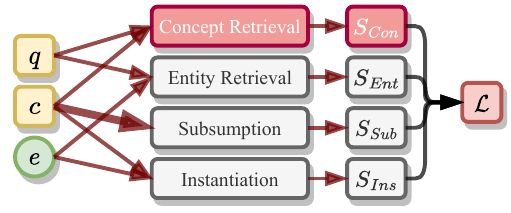}
  \vspace{-0.21cm}
  \caption{Illustration of obtaining $\mathcal{L}$ for optimization.}
  \label{fig:loss}
  \vspace{-0.21cm}
\end{figure}
\label{train}

The parameters to optimize in our model TAR include 
the entity embedding matrix $\mathbf{E}_e$, the concept embedding matrix $\mathbf{E}_c$ for
the \textit{basic} representation of concepts that is out of domain
$\Delta^\mathcal{I}$, the relation embedding matrix $\mathbf{E}_r$, $\theta$ in Section \ref{subsumption}, and $\Omega$ in Section \ref{sec:e-retrieval}. 
In the training stage, we sample $m$ negative samples for each
positive instance of concept-level answering $[q](c^+)$ by corrupting
$c^+$ with randomly sampled $c^-_i \in \mathcal{C}$
($i = {1, \cdots, m}$). Similarly, negative samples for entity-level answering
$[q](e^-_i)$ are obtained by corrupting $e^+$ in $[q](e^+)$ with
randomly sampled $e^-_i \in \mathcal{E}$. For subsumption and
instantiation, both sides of the concept-concept pairs and
concept-entity pairs are randomly corrupted following the same
procedure.
The loss of TAR is defined as
\begin{equation}
  \mathcal{L} = - \frac{1}{4m} \sum_{n \in N} \sum_{i=1}^{m} \log \sigma (S_n^+ - S_{n_i}^-)
\label{eqn:loss}
\end{equation}
where $N = \{ Con, Ent, Sub, Ins \}$ denotes the set of the four
included task discussed Section \ref{operator}, $S_n^+$ (or $S_{n_i}^-$)
denotes the predicted similarity or degree of inclusion of the
positive (or negative) sample according to task $n$. 
The overall
optimization process of $\mathcal{L}$  is outlined in Algorithm 1 in supplementary materials.

In the inference stage, we predict $S_{Con}$ (or $S_{Ent}$) for every
candidate concept $c \in \mathcal{C}$ (or entity $e \in \mathcal{E}$)
regarding to query $q$ and select the top-$k$ results to be the concept-level TBox answers $\{ a_c \}$ (or
entity-level ABox answers $\{ a_e \}$) for query $q$. Thus, we are able
to achieve TA-NLR by providing the comprehensive answers
$\{a\} = \{a_e\} \bigcup \{a_c\}$. Although subsumption in Section
\ref{subsumption} and instantiation in Section \ref{instantiation} are
not included in the inference stage, they empowered TAR to better
represent and operate concepts by providing training
instances and extra supervision signals.
\begin{table*}[t]
\small
    \renewcommand\arraystretch{0.9}
  	\renewcommand\tabcolsep{3pt}
  \centering
  \caption{Performance of providing concept-level TBox answers. The best results are in boldface.}
  \vspace{-0.25cm}
    \begin{tabular}{c||c||ccccc||cccc||c||ccccc||cccc||c}
    \toprule
    \multicolumn{1}{c}{} &       & \multicolumn{10}{c||}{MRR}                                                    & \multicolumn{10}{c}{Hit@3} \\
    
\cmidrule{3-22}    \multicolumn{1}{c}{} &       & 1p    & 2p    & 3p    & 2i    & 3i    & pi    & ip    & 2u    & up    & avg   & 1p    & 2p    & 3p    & 2i    & 3i    & pi    & ip    & 2u    & up    & avg \\
\midrule
    \multirow{4}[0]{*}{YAGO4} & GQE \cite{hamilton2018embedding}   & 35.3  & 49.5  & 33.7  & 51.3  & 43.9  & 11.8  & 9.1   & 14.4  & 6.6   & 28.4  & 43.7  & 67.4  & 44.9  & 67.1  & 49.9  & 15.0  & 12.3  & 19.7  & 9.4   & 36.6 \\
          & Q2B \cite{ren2019query2box}   & 37.3  & 53.4  & 59.6  & 55.0  & 47.6  & 2.1   & 1.6   & 1.7   & 1.1   & 28.8  & 47.1  & 75.4  & 78.2  & \textbf{70.0} & 58.2  & 2.8   & 1.9   & 1.9   & 1.4   & 37.4 \\
          & BetaE \cite{ren2020beta} & 39.0  & 57.0  & 58.9  & 52.9  & 45.8  & 10.4  & 8.9   & 2.7   & 6.5   & 31.3  & 47.8  & 74.7  & 76.7  & 64.2  & 52.5  & 10.1  & 10.2  & 2.9   & 5.0   & 38.2 \\
          & FuzzQE \cite{chen2022fuzzy} & 
          34.1&	50.3&	44.4&	52.9&	44.0&	13.2&	11.7&	19.0&	8.3&	30.9&	41.6&	69.8&	59.4&	67.5&	52.8&	16.9&	13.3&	23.4&	10.4&	39.5\\
          & \textbf{TAR} & \textbf{51.3} & \textbf{76.4} & \textbf{82.7} & \textbf{55.9} & \textbf{53.9} & \textbf{51.3} & \textbf{48.9} & \textbf{54.3} & \textbf{45.9} & \textbf{57.8} & \textbf{60.3} & \textbf{88.8} & \textbf{88.7} & 66.4  & \textbf{65.5} & \textbf{60.2} & \textbf{57.1} & \textbf{59.9} & \textbf{52.5} & \textbf{66.6} \\
          \midrule
    \multirow{4}[1]{*}{DBpedia} & GQE \cite{hamilton2018embedding}   & 27.1  & 35.5  & 32.5  & 30.5  & 32.0  & 14.0  & 14.7  & 9.8   & 11.8  & 23.1  & 28.7  & 42.9  & 40.0  & 32.5  & 36.0  & 13.0  & 14.4  & 7.2   & 10.7  & 25.0 \\
          & Q2B \cite{ren2019query2box}   & 26.4  & 35.7  & 32.6  & 30.4  & 29.9  & 13.5  & 14.4  & 10.3  & 11.2  & 22.7  & 28.2  & 41.4  & 38.0  & 32.7  & 33.5  & 11.7  & 13.0  & 8.4   & 9.3   & 24.0 \\
          & BetaE \cite{ren2020beta} & 30.4  & 38.7  & 40.0  & 32.9  & 34.2  & 14.8  & 11.4  & 5.6   & 9.0   & 24.1  & 34.1  & 45.4  & 50.8  & 37.2  & 41.2  & 14.6  & 10.9  & 4.0   & 8.0   & 27.4 \\
          & FuzzQE \cite{chen2022fuzzy} &
          26.7&	34.7&	32.1&	28.1&	28.4&	16.9&	16.5&	13.1&	14.6&	23.5&	29.0&	39.2&	38.2&	30.5&	29.7&	15.6&	15.9&	11.2&	12.7&	24.7\\
          & \textbf{TAR} & \textbf{55.0} & \textbf{80.8} & \textbf{80.7} & \textbf{42.9} & \textbf{36.7} & \textbf{42.0} & \textbf{28.5} & \textbf{63.4} & \textbf{64.9} & \textbf{55.0} & \textbf{62.4} & \textbf{83.7} & \textbf{83.6} & \textbf{50.9} & \textbf{43.7} & \textbf{45.0} & \textbf{29.9} & \textbf{67.2} & \textbf{67.3} & \textbf{59.3} \\
    \bottomrule
    \end{tabular}%
      \vspace{-0.1cm}
  \label{abductiveresults}%
\end{table*}%

\section{Experiments}
We conduct extensive experiments to answer the
following research questions: 
\textbf{RQ1} How to properly compare TAR with methods that do not give concept-level TBox answers? 
\textbf{RQ2} How does
TAR perform for providing concept-level TBox answers?  
\textbf{RQ3} How does TAR
perform for providing entity-level ABox answers? 
\textbf{RQ4} How do the introduced
subsumption and instantiation operators affect the performance of TAR?

\subsection{Experimental Settings}

\subsubsection{\textbf{Baselines (RQ1)}}
\label{onemorehop}
The considered baseline methods are the three most established methods
in NLR, namely GQE \cite{hamilton2018embedding}, Q2B
\cite{ren2019query2box}, and BetaE \cite{ren2020beta}, along with a recent method FuzzQE \cite{chen2022fuzzy} that applies fuzzy operations in a different way.
Since the regular neural logical reasoners can only
provide entity-level answers, we
need to come up with a way to make them give concept-level answers, so as to be
compared with our proposed TAR on concept-level reasoning.

Therefore, we introduce the \textit{One-more-hop} experiment.
That is, we exploit all the information given by
$\mathcal{KB}= (\mathcal{T}, \mathcal{A})$ and simply degrade concepts
to entities in the training stage.  Specifically, we first augment
$\mathcal{A}_{ec}$ by the transductive links provided by
$\mathcal{T}$. Then we combine the augmented $\mathcal{A}_{ec}$ and
$\mathcal{A}_{ee}$ to form the new knowledge graph
$\mathcal{KG}^\prime$. Note that part of the entities in
$\mathcal{KG}^\prime$ are the degraded concepts and
$\mathcal{KG}^\prime$ contains an additional relation $r_{ec}$ to
describe the \textit{isInstanceOf} relationship between an entity and
a concept. Thus, we construct training examples of various types of
queries using $\mathcal{KG}^\prime$ and update model parameters following \cite{ren2019query2box}.

In the inference stage, two sets of candidate entities are prepared
for each query. The first is the regular entity-level candidate set, which can be ranked following the original
papers \cite{hamilton2018embedding,ren2019query2box,ren2020beta}. Another set contains the degraded concepts. To
predict the possibility of a concept being an answer of a
query $[q](?)$, we add one more projection operation with the relation
$r_{ec}$, so as to construct the query:
$[q^\prime](?) = [q \xrightarrow{r_{ec}}](?)$. In other words,
concept-level reasoning is implicitly achieved by an additional hop asking
the \textit{isInstanceOf} upon entity retrieval queries, i.e., the \textit{One-more-hop}.

\subsubsection{\textbf{Datasets}}
We conduct experiments on two commonly-used real-world large-scale
knowledge bases, namely
YAGO4 and
DBpedia.
Specifically, we use English Wikipedia version\footnote{https://yago-knowledge.org/downloads/yago-4} of YAGO4 and 2016-10 release\footnote{http://downloads.dbpedia.org/wiki-archive/downloads-2016-10.html} of DBpedia.
To preprocess the dataset for the TA-NLR problem, we first filter out
low-degree entities in $\mathcal{A}_{ee}$ and $\mathcal{A}_{ec}$ with
the threshold 5. Then we split $\mathcal{A}_{ee}$ to two sets with the
ratio 95\% and 5\% for training and evaluation, respectively. 
We use the same procedure as BetaE \cite{ren2020beta} to construct instances of logical queries from $\mathcal{A}_{ee} = \mathcal{KG}$. 
We use
all the triples in $\mathcal{A}_{ee}$ in the training set as training
examples of {\em 1p} queries and randomly select certain amount of
training and evaluation examples for each of the other types of
queries as stated in Table \ref{statistic}. 
We then split the evaluation set of each type of queries to
the validation set and the testing set.  We summarize the statistics
of datasets in Table \ref{statistic}.

\begin{table*}[t]
\small
    \renewcommand\arraystretch{0.9}
  	\renewcommand\tabcolsep{3pt}
  \centering
  \caption{Performance of providing entity-level ABox answers. The best results are in boldface.}
    \vspace{-0.21cm}
    \begin{tabular}{c||c||ccccc||cccc||c||ccccc||cccc||c}
    \toprule
    \multicolumn{1}{c}{\textcolor[rgb]{ .733,  .733,  .733}{}} &       & \multicolumn{10}{c||}{MRR}                                                    & \multicolumn{10}{c}{Hit@3} \\
\cmidrule{3-22}    \multicolumn{1}{c}{\textcolor[rgb]{ .733,  .733,  .733}{}} &       & 1p    & 2p    & 3p    & 2i    & 3i    & pi    & ip    & 2u    & up    & avg   & 1p    & 2p    & 3p    & 2i    & 3i    & pi    & ip    & 2u    & up    & avg \\
\midrule
    \multirow{4}[0]{*}{YAGO4} & GQE \cite{hamilton2018embedding}   & 25.8  & 15.8  & 4.6   & 26.4  & 28.4  & 21.7  & 18.1  & 9.6   & 18.2  & 18.7  & 29.7  & 17.3  & 4.6   & 29.0  & 31.4  & 23.5  & 18.6  & 13.6  & 17.4  & 20.6 \\
          & Q2B \cite{ren2019query2box}   & 24.5  & 17.2  & 6.8   & 25.7  & 28.8  & 24.5  & 20.2  & 8.6   & 17.7  & 19.3  & 28.4  & 20.0  & 7.0   & 29.5  & 34.4  & 27.0  & 21.3  & 11.6  & 18.5  & 22.0 \\
          & BetaE \cite{ren2020beta} & 28.2  & 19.7  & 9.5   & 30.1  & 33.4  & 28.3  & 22.0  & 10.6  & 20.7  & 22.5  & 31.4  & 22.3  & 12.0  & 33.5  & 37.5  & 31.3  & 24.1  & 12.5  & \textbf{23.8} & 25.4 \\
        & FuzzQE \cite{chen2022fuzzy} & 
        26.4&	15.6&	4.0&	26.8&	28.3&	21.2&	18.0&	10.4&	18.1&	18.8&	30.7&	17.3&	4.1&	30.0&	32.3&	22.9&	18.1&	12.5&	18.6&	20.7\\
          & \textbf{TAR} & \textbf{34.9} & \textbf{23.8} & \textbf{15.1} & \textbf{50.9} & \textbf{61.6} & \textbf{31.6} & \textbf{35.3} & \textbf{13.9} & \textbf{21.3} & \textbf{32.0} & \textbf{39.6} & \textbf{27.3} & \textbf{18.4} & \textbf{56.6} & \textbf{70.2} & \textbf{34.2} & \textbf{39.1} & \textbf{14.9} & 23.7  & \textbf{36.0} \\
          \midrule
    \multirow{4}[1]{*}{DBpedia} & GQE \cite{hamilton2018embedding}   & 19.6  & 16.1  & 18.2  & 31.4  & 38.2  & 16.8  & 28.5  & 10.1  & 18.0  & 21.9  & 25.2  & 18.9  & 19.8  & 26.5  & 43.6  & 17.2  & 33.3  & 12.6  & 21.1  & 24.2 \\
          & Q2B \cite{ren2019query2box}   & 16.3  & 13.7  & 15.4  & 22.6  & 28.5  & 18.1  & 22.8  & 7.4   & 12.9  & 17.5  & 21.3  & 16.5  & 17.7  & 27.5  & 31.9  & 19.7  & 25.7  & 9.1   & 15.4  & 20.7 \\
          & BetaE \cite{ren2020beta} & 20.2  & 20.1  & 19.3  & 25.4  & 29.7  & 24.2 & 25.2  & 12.3  & \textbf{24.2} & 22.3  & 20.9  & 23.2  & 21.7  & 27.5  & 32.9  & 27.5  & 28.4  & 13.0  & \textbf{25.1}  & 24.5 \\
        & FuzzQE \cite{chen2022fuzzy} &
        15.1&	14.1&	16.3&	22.3&	29.1&	\textbf{27.6}&	23.1&	6.8&	13.7&	18.7&	18.5&	16.3&	18.5&	26.5&	33.0&	\textbf{32.6}&	25.4&	8.0&	16.0&	21.6\\
          & \textbf{TAR} & \textbf{28.8} & \textbf{24.5} & \textbf{24.4} & \textbf{38.4} & \textbf{46.3} & 20.1  & \textbf{33.6} & \textbf{14.0} & 20.6  & \textbf{27.9} & \textbf{34.6} & \textbf{28.0} & \textbf{29.0} & \textbf{44.6} & \textbf{54.8} & 21.4 & \textbf{40.6} & \textbf{17.8} & 23.0 & \textbf{32.6} \\
    \bottomrule
    \end{tabular}%
    \vspace{-0.1cm}
  \label{inductiveresults}%
\end{table*}%

\subsubsection{\textbf{Implementation Details}}
\label{inplementation}
We implement TAR using
PyTorch and conduct all the experiments with Nvidia RTX 3090 GPUs and Intel Xeon CPUs.  In
the training stage, the initial learning rate of the Adam
\cite{kingma2014adam} optimizer, the embedding dimension $d$, and the
batch size, are tuned by grid searching within \{$1e^{-2}$, $1e^{-3}$,
$1e^{-4}$, $1e^{-5}$\}, \{128, 256, 512\}, and \{256, 512, 1024\}, respectively.
We keep the number of corrupted negative samples for each positive
sample $m$, the small value $\epsilon$, the exponent value $p$, the margin $\gamma$, and
the adopted type of $t$-norm as 4, $1e^{-12}$, 1, 12, and
$\top_{\text {prod }}$, respectively. We employ early stop with validation
interval of 50 and tolerance of 3 for model training.  In the test
phase, following \cite{ren2019query2box}, we use the filtered setting
and report the averaged results of Mean Reciprocal Rank (MRR) and
Hits@3 over 3 independent runs.

\subsection{Concept-level TBox Answers (RQ2)}
We conduct the \textit{One-more-hop} experiment as described in
Section \ref{onemorehop} to answer RQ2. As shown in
Table.\ref{abductiveresults}, our proposed TAR consistently
outperforms baseline methods on various evaluation metric with large
margins. For the \textit{basic} queries summarized in
Figure \ref{fig:types} that are simply projections and intersections,
our proposed TAR significantly improved the performance of providing concept-level TBox answers, especially for the multiple projection queries {\em 1p}, {\em 2p},
and {\em 3p}. For \textit{extra} queries in Figure.\ref{fig:types} that
are more complex in terms of including unions or combined logical
operations, we even boosted the performance exponentially. The average
performance of TAR is also significantly better than baseline
methods.

The superior performance of TAR can be explained in two folds. First, due to the lack of reasoning capabilities across TBox and ABox, GQE, Q2B, and BetaE need to do
reasoning over more complicated queries. For example, baseline methods
need to do reasoning over an {\em ipp} query
$[((h_1 \xrightarrow{r_1} ) \wedge (h_2 \xrightarrow{r_2}))
\xrightarrow{r_{ec}}](?)$ to provide concept-level answers
of an {\em ip} query
$[(h_1 \xrightarrow{r_1} ) \wedge \neg(h_2 \xrightarrow{r_2})
](?)$. Therefore, {\em 1p} queries become {\em 2p} queries for baseline
methods, {\em 2p} becomes {\em 3p}, and so on.  Thus, the complexity of the
transformed queries limits the baseline performance.  Second, explicit
supervision signals for concept-level reasoning are not provided by the
baselines. That is to say, since the concepts are degraded as entities, regular NLR methods could not explicitly feed the empirical error on concept-level answers back to update the model parameters.
It is thus understandable that the baseline methods cannot perform well  to provide concept-level TBox answers,
especially on \textit{extra} queries that are more complicated and require
supervision signals more eagerly.

\subsection{Entity-level ABox Answers (RQ3)}
Although TAR is designed for TA-NLR, it is interesting to know the performance of TAR on entity-level reasoning only (for answering  RQ3). 
The results in Table \ref{inductiveresults} show that TAR also outperforms regular NLR methods on most types of queries on various metrics.
The performance gain of TAR should be credited to its capability of 
representing and operating on concepts. It thus has additional information of the relationships among queries, entities, and concepts, which are helpful for providing entity-level Abox answers. 

\begin{table}[t!]
  \centering
  \caption{Ablation Study on Subsumptions and Instantiation upon DBpedia dataset. The best MRR results are in boldface. }
  \vspace{-0.3cm}
  	\renewcommand\tabcolsep{1.1pt}
    \begin{tabular}{c||ccccc|cccc||c}
    \toprule
    \textbf{TBox answers} & 1p    & 2p    & 3p    & 2i    & 3i    & pi    & ip    & 2u    & up    & avg \\
    \midrule
     w/o Sub & 53.3  & 72.4  & 71.5  & 19.0  & 15.8  & 24.1  & 19.3  & 59.9  & 61.7  & 44.1 \\
     w/o Ins & 52.8  & 68.8  & 67.7  & 38.3  & 35.4  & 34.2  & 19.3  & 56.2  & 61.6  & 48.3 \\
    \textbf{TAR} & \textbf{55.0} & \textbf{80.8} & \textbf{80.7} & \textbf{42.9} & \textbf{36.7} & \textbf{42.0} & \textbf{28.5} & \textbf{63.4} & \textbf{64.9} & \textbf{55.0} \\
    \midrule
    \textbf{ABox answers} & 1p    & 2p    & 3p    & 2i    & 3i    & pi    & ip    & 2u    & up    & avg \\
    \midrule
     w/o Sub & 17.7  & 20.1  & 21.0  & 18.0  & 18.8  & 14.9  & 22.7  & 9.1   & 16.0  & 17.6 \\
     w/o Ins & 17.4  & 18.8  & 19.0  & 18.1  & 18.7  & 14.3  & 22.9  & 8.7   & 17.7  & 17.3 \\
    \textbf{TAR} & \textbf{28.8} & \textbf{24.5} & \textbf{24.4} & \textbf{38.4} & \textbf{46.3} & \textbf{20.1} & \textbf{33.6} & \textbf{14.0} & \textbf{20.6} & \textbf{27.9} \\
    \bottomrule
    \end{tabular}%
  \label{tab:ablation}%
  \vspace{-0.3cm}
\end{table}%

\subsection{Ablation Study (RQ4)}
We conduct ablation study on DBpedia dataset
to answer RQ4. As shown in Table \ref{tab:ablation}, when the \textit{Subsumption} task is not included, i.e., ontological axioms in $\mathcal{T}$ are not used and $S_{Sub}$ is not computed, TAR \textit{w/o Sub} underperforms TAR on all types of queries for both tasks.  Such results clearly demonstrate the importance of the relational information between concepts to be used for TA-NLR, and the effectiveness of the designed operator in Section \ref{subsumption} for handling such information. On the other hand, TAR consistently outperforms \textit{w/o Ins} on all types of queries. This verifies   that the relational information about $isInstanceOf$ in $\mathcal{A}_{ec}$ is vital for TA-NLR, and the pair-wise degraded fuzzy set generation process introduced in Section \ref{instantiation} is effective to tackle with \textit{Instantiation}. More experimental results about ablation study are reported in Table 1 in supplementary materials.

\subsection{Case Study of Concept Representation}
\begin{table}[t!]
      \renewcommand\arraystretch{0.9}
  \centering
  \caption{Top 10 entities with   highest degrees of membership in concept ``\emph{place}'' $c=$ <http://dbpedia.org/ontology/Place>. The  real-world places are in bold-face.}
  \vspace{-0.2cm}
  \small
    \begin{tabular}{c||c||c}
    \toprule
    Entity $e$ & Info  & $\mu(e)$ \\
    \midrule
    \textbf{Province\_of\_L'Aquila} & \textbf{A province of Italy} & 1.0000 \\
    Lietuvos\_krepšinio\_lyga      & \textcolor[rgb]{ .125,  .129,  .133}{A sport league in Lithuania} & 0.9988 \\
    Moghreb\_Tétouan      & A sport team in Morocco & 0.9988 \\
    \textbf{Quebec\_Route\_132} & \textbf{A highway in Canada} & 0.9986 \\
    \textbf{School\_of\_Visual\_Arts} & \textbf{A college in New York} & 0.9968 \\
    League\_of\_Ireland & \textcolor[rgb]{ .125,  .129,  .133}{A sport league in Ireland} & 0.9777 \\
    \textbf{Pančevo}      & \textbf{A city in Serbia} & 0.9762 \\
    \textbf{Kolar} & \textbf{A city in India} & 0.9755 \\
    Kemco & A company in Japan& 0.9740 \\
    \textbf{Deyr\_County} & \textbf{A county in Iran} & 0.9740 \\
    \bottomrule
    \end{tabular}%
  \label{tab:case}%
  \vspace{-0.2cm}
\end{table}%

To gain insights of the fuzzy set representation of concept learned in TAR,  we present a case study on the concept  ``\emph{place}''. 
Table \ref{tab:case} shows the top-10 entities $e$ ranked by their degrees of membership $\mu(\cdot)$ to $c =$ <http://dbpedia.org/ontology/Place>.
The entities of real-world places are highlighted in boldface. As 6 out of the top 10 entities 
are correct (real places), we believe fuzzy sets are capable of representing concepts in domain $\Delta^{\mathcal{I}}$ as vague sets of entities. Regarding to the other 4 entities, we observe that they are representative sport teams, leagues, or companies of a corresponding  region. Although they are not real places, it makes sense that they have high degrees of membership $\mu(\cdot)$ to $c$,  as they are strongly associated to the corresponding places. Therefore, the results demonstrate that the fuzzy set based TAR is capable of take the advantage of vagueness to explore the highly related entities.

\section{Conclusion}
In conclusion, we formulated the TA-NLR problem that performs neural logical reasoning across TBox and ABox. This is a novel problem and  of great importance for users, downstream tasks, and ontological applications. The key challenges for addressing TA-NLR
are the incorporation of concepts, representation of concepts, and operator on concepts. Accordingly, we propose TAR that properly incorporates ontological axioms, represents concepts and queries as fuzzy sets, and operates on concepts based on fuzzy sets. Extensive experimental results demonstrate the effectiveness of TAR for TA-NLR. The processed datasets and code are ready to be published to foster further research of TA-NLR.

\clearpage

\bibliographystyle{ACM-Reference-Format} \bibliography{references}

\clearpage
\appendix
\section{Supplementary Materials}

\subsection*{Details of Entity Retrieval}
\label{inductive}
Here, we elaborate the method to obtain the query embedding $\mathbf{q}$ for providing entity-level ABox answers, i.e., $f(\cdot)$ with parameters $\Omega$. 
We use the integrated implementation\footnote{https://github.com/snap-stanford/KGReasoning} to obtain $\mathbf{q}$. Specifically, the \textbf{projection} operation $\mathbf{x} \xrightarrow{r}$ that project an entity or query embedding $\mathbf{x}$ with relation $r$ is resolved by:
\begin{equation*}
    \mathbf{q} = \mathbf{x} + \mathbf{r},
\end{equation*}
where $\mathbf{x} \in \mathbb{R}^d$ is another query embedding that is obtained in advance or an entity embedding obtained by looking up $\mathbf{E}_e \in \mathbb{R}^{|\mathcal{E}| \times d}$ by rows. The \textbf{intersection} of two query embeddings $\mathbf{q_1}$ and $\mathbf{q_2}$ is resolved by:
\begin{equation*}
    \mathbf{q} = a(\mathbf{q_1} \oplus \mathbf{q_2}; \Omega)_1 * \mathbf{q_1} + a(\mathbf{q_1} \oplus \mathbf{q_2}; \Omega)_2 * \mathbf{q_2},
\end{equation*}
where $\oplus$ denotes matrix concatenation over the last dimension, $\Omega$ denote s the parameters of $a(\cdot)$, and $a(\cdot)$ is a two-layer feed-forward network with $Relu$ activation. $a(\cdot)_1$ and $a(\cdot)_2$ represent the first and second $d$ attention weights, respectively. The \textbf{union} of two query embeddings $\mathbf{q_1}$ and $\mathbf{q_2}$ is resolved by:
\begin{equation*}
    \mathbf{q} = \max(\mathbf{q_1}, \mathbf{q_2})_{-1},
\end{equation*}
where $\max(\cdot)_{-1}$ denotes the max operation over the last dimension.

\subsection*{Additional Experiments}
\begin{table}[htbp]
  \centering
  \vspace{-0.35cm}
    \caption*{Table 1: Ablation Study on Subsumptions and Instantiation upon DBpedia dataset. The best Hit@3 results are in boldface. }
  \small
  	\renewcommand\tabcolsep{1.6pt}
    \begin{tabular}{c||ccccc||cccc||c}
    \toprule
    \textbf{TBox answers} & 1p    & 2p    & 3p    & 2i    & 3i    & pi    & ip    & 2u    & up    & avg \\
    \midrule
    w/o CC & 61.4  & 73.8  & 71.8  & 22.9  & 20.1  & 28.8  & 21.6  & \textbf{67.5} & 62.6  & 47.8 \\
    w/o EC & 58.9  & 69.7  & 68.0  & 42.8  & 39.6  & 41.2  & 22.0  & 66.4  & 62.9  & 52.4 \\
    \textbf{TAR} & \textbf{62.4} & \textbf{83.7} & \textbf{83.6} & \textbf{50.9} & \textbf{43.7} & \textbf{45.0} & \textbf{29.9} & 67.2  & \textbf{67.3} & \textbf{59.3} \\
    \midrule
    \textbf{ABox answers} & 1p    & 2p    & 3p    & 2i    & 3i    & pi    & ip    & 2u    & up    & avg \\
    \midrule
    w/o CC & 20.5  & 22.8  & 23.0  & 18.8  & 20.6  & 14.7  & 25.6  & 10.4  & 18.4  & 19.4 \\
    w/o EC & 19.4  & 19.1  & 20.5  & 19.1  & 20.1  & 15.0  & 25.7  & 9.1   & 20.3  & 18.7 \\
    \textbf{TAR} & \textbf{34.6} & \textbf{28.0} & \textbf{29.0} & \textbf{44.6} & \textbf{54.8} & \textbf{21.4} & \textbf{40.6} & \textbf{17.8} & \textbf{23.0} & \textbf{32.6} \\
    \bottomrule
    \end{tabular}%
  \label{hrablation}%
\end{table}%

\begin{breakablealgorithm}
\caption{The learning procedure of \textbf{TAR}.}
\label{alg:algorithm}
\begin{algorithmic}[1] 
\REQUIRE An ontological knowledge base $\mathcal{KB} = (\mathcal{T}, \{\mathcal{A}_{ee}, \mathcal{A}_{ec}\}) $. \\
$\mathcal{E}$ denotes the set of entities; \\
$\mathcal{C}$ denotes the set of concept (names) ; \\
$\mathcal{R}$ denotes the set of relations ; \\
\ENSURE $\mathbf{E}_e$ denotes the entity embedding matrix;\\
$\mathbf{E}_c$ denotes the concept embedding matrix;\\
$\mathbf{E}_r$ denotes the relation embedding matrix;\\
$\Theta$ denotes the parameters of $h(\cdot)$; \\
$\Omega$ denotes the parameters of $f(\cdot)$. 
\STATE // Start training. 
\STATE Initialize $\mathbf{E}_e$, $\mathbf{E}_c$, $\mathbf{E}_r$, $\Theta$ and $\Omega$.
\FOR{each training episode}
    \STATE // \textbf{Concept Retrieval}.
        \FOR{each query $[q](?)$}
            \STATE Sample a concept-level answer $c^+ \in \mathcal{C}$ as a positive instance and $m$ non-answer concepts $\{c^-_1, \cdots, c^-_m\}$ as negative instances;
            \STATE Represent $c^+$ as $FS_{c^+}$ by Eq.(6);
            \STATE Represent $q$ as $FS_{q}$ by Eq.(8), (9), (10), and (11);
            \STATE Calculate $S_{Con}^+$ by Eq.(12) and (13) given $FS_{c^+}$ and $FS_{q}$;
            \FOR{each negative instance $c^-_i$}
                \STATE Represent $c^-_i$ as $FS_{c^-_i}$ by Eq.(6);
                \STATE Calculate $S_{Con_i}^-$ by Eq.(12) and (13) given $FS_{c^-_i}$ and $FS_{q}$;
            \ENDFOR
        \ENDFOR
    \STATE // \textbf{Entity Retrieval}.
        \FOR{each query $[q](?)$}
            \STATE Sample an entity-level answer $e^+ \in \mathcal{E}$ as a positive instance and $m$ non-answer entities $\{e^-_1, \cdots, e^-_m\}$ as negative instances;
            \STATE $\mathbf{e}^+$ $\leftarrow$ Look up $\mathbf{E}_e$ by rows; $\mathbf{q}$ $\leftarrow$ $f(q;\Omega)$;
            \STATE Calculate $S_{Ent}^+$ by Eq.(14) given $\mathbf{e}^+$ and $\mathbf{q}$;
            \FOR{each negative instance $e^-_i$}
                \STATE $\mathbf{e}^-_i$ $\leftarrow$ Look up $\mathbf{E}_e$ by rows;
                \STATE Calculate $S_{Ent_i}^-$ by Eq.(14) given $\mathbf{e}^-_i$ and $\mathbf{q}$;
            \ENDFOR
        \ENDFOR
    \STATE // \textbf{Subsumption}.
        \FOR{each pair of concepts $(c_1, c_2)$}
            \STATE Sample $m$ concepts $\{c^-_1, \cdots, c^-_m\}$ as negative instances;
            \STATE $\mathbf{c}_1$, $\mathbf{c}_2$ $\leftarrow$ Look up $\mathbf{E}_c$ by rows;
            \STATE Calculate $S_{Sub}^+$ by Eq.(15) given $\mathbf{c}_1$ and $\mathbf{c}_2$;
            \FOR{each negative instance $c^-_i$}
                \STATE $\mathbf{c}^-_i$ $\leftarrow$ Look up $\mathbf{E}_c$ by rows;
                \STATE Calculate $S_{Sub_i}^-$ by Eq.(15) given $(\mathbf{c}_1, \mathbf{c}^-_i)$ or $(\mathbf{c}^-_i, \mathbf{c}_2)$ with equal probability;
            \ENDFOR
        \ENDFOR
    \STATE // \textbf{Instantiation}.
        \FOR{each pair of concept and entity $(c, e)$}
            \STATE Sample $\frac{m}{2}$ negative concepts $\{c^-_1, \cdots, c^-_{\frac{m}{2}}\}$;
            \STATE Sample $\frac{m}{2}$ negative entities $\{e^-_{\frac{m}{2} + 1}, \cdots, e^-_m\}$;
            \STATE $\mathbf{c}$ $\leftarrow$ Look up $\mathbf{E}_c$ by rows;
            \STATE $\mathbf{e}$ $\leftarrow$ Look up $\mathbf{E}_e$ by rows;
            \STATE Calculate $S_{Ins}^+$ by Eq.(16) given $\mathbf{c}$ and $\mathbf{e}$;
            \FOR{each negative concept $c^-_i$}
                \STATE $\mathbf{c}^-_i$ $\leftarrow$ Look up $\mathbf{E}_c$ by rows;
                \STATE Calculate $S_{Ins_i}^-$ by Eq.(16) given $(\mathbf{c}^-_i, \mathbf{e})$;
            \ENDFOR
            \FOR{each negative entity $e^-_i$}
                \STATE $\mathbf{e}^-_i$ $\leftarrow$ Look up $\mathbf{E}_e$ by rows;
                \STATE Calculate $S_{Ins_i}^-$ by Eq.(16) given $(\mathbf{c}, \mathbf{e}^-_i)$;
            \ENDFOR
        \ENDFOR
    \STATE Calculate $\mathcal{L}$ by Eq.(17);
    \STATE Update $\mathbf{E}_e \leftarrow \partial \mathcal{L} / \partial \mathbf{E}_e$;
    Update $\mathbf{E}_c \leftarrow \partial \mathcal{L} / \partial \mathbf{E}_c$;
    Update $\mathbf{E}_r \leftarrow \partial \mathcal{L} / \partial \mathbf{E}_r$;
    Update $\Theta \leftarrow \partial \mathcal{L} / \partial \Theta$;
    Update $\Omega \leftarrow \partial \mathcal{L} / \partial \Omega$;
\ENDFOR
\STATE \textbf{return updated $\mathbf{E}_e$, $\mathbf{E}_c$, $\mathbf{E}_r$, $\Theta$ and $\Omega$.} 
\end{algorithmic}
\end{breakablealgorithm}

\end{document}